\documentclass[a4paper]{article}
\usepackage{iwslt17,amssymb,amsmath,epsfig}
\usepackage{color}
\usepackage{verbatim}
\usepackage[utf8]{inputenc}
\usepackage[T1]{fontenc}
\def\reg{{\rm\ooalign{\hfil
     \raise.07ex\hbox{\scriptsize R}\hfil\crcr\mathhexbox20D}}}

\newcommand\sml[1]{\textcolor{black}{#1}}

\title{Improving Zero-Shot Translation of Low-Resource Languages}

 \makeatletter
 \def\name#1{\gdef\@name{#1\\}}
 \makeatother
\name{{\em Surafel M. Lakew$^{1,2}$, Quintino F. Lotito $^{2,*}$\thanks{* Work done during a summer internship at FBK.}, Matteo Negri$^{1}$, Marco Turchi$^{1}$, Marcello Federico$^{1}$}}

\address{$^{1}$Fondazione Bruno Kessler, Trento, Italy  \\
$^{2}$University of Trento, Trento, Italy \\
{\small \tt \{lakew,lotito,negri,turchi,federico\}@fbk.eu}
}
\begin{document}
\maketitle

\begin{abstract}
Recent work on multilingual neural machine translation reported competitive 
performance with respect to bilingual models and surprisingly good performance 
even on (zero-shot) translation directions not observed at training time. 
We investigate here a zero-shot translation in a particularly low-resource 
multilingual setting. We propose a simple iterative training procedure that
leverages a duality of translations directly generated by the system for the zero-shot directions.
The translations produced by the system (sub-optimal since they contain mixed language from the shared vocabulary), are then used together with the original parallel data to feed and iteratively re-train the multilingual network. 
Over time, this allows the system to learn from its own generated and increasingly better output.
Our approach shows to be effective in improving the two zero-shot directions of our multilingual model. In particular, we observed gains of about $9$ BLEU points over a baseline multilingual model and up to $2.08$ BLEU  over a pivoting mechanism using two bilingual models. Further analysis shows that there is also a slight improvement in the non-zero-shot language directions.
\end{abstract}

\section{Introduction}
Machine translation of low-resource languages represents a challenge for neural machine translation (NMT) \cite{koehn2017six}. Recent efforts in multilingual NMT (Multi-NMT) \cite{johnson2016google,ha2016toward} have shown to improve translation performance in low-resource settings. Multi-NMT models can be trained with parallel corpora of several language pairs to work in \textit{many-to-one, one-to-many, or many-to-many} translation directions. A simple approach, named \textit{target-forcing}~\cite{ha2016toward}, is to prepend to the source sentence a tag specifying the target language, both at training and testing time. In addition to performance gains for low-resource languages, the benefit of Multi-NMT is the possibility to perform zero-shot translation, i.e. across directions that were not observed at training time. 

\begin{figure}[!t]
\centerline{\epsfig{figure=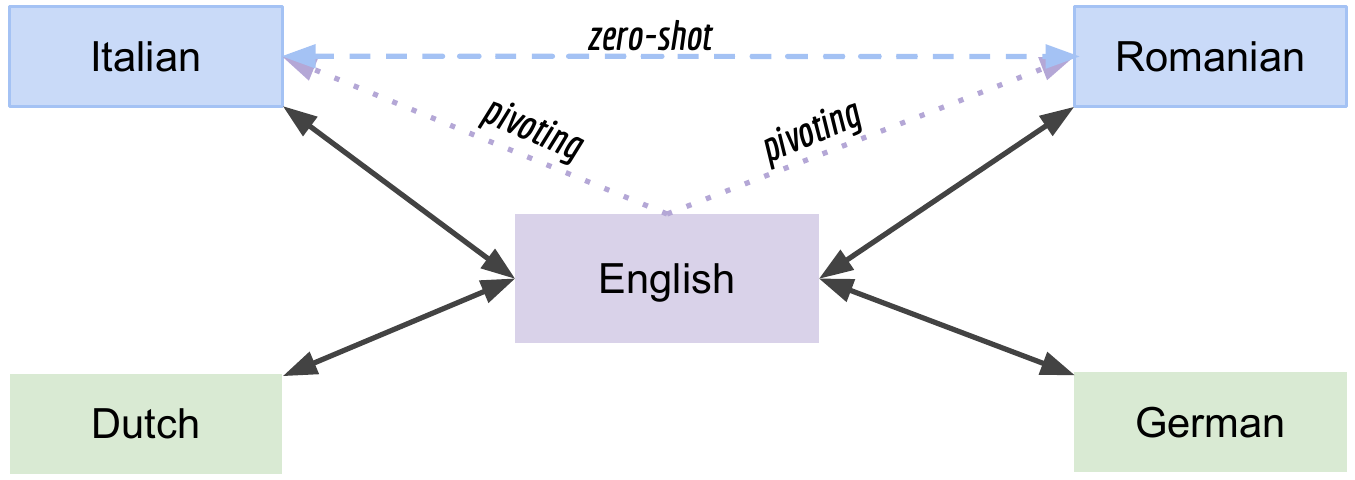,width=3.4in}}
\caption{Our zero-shot translation setting for Italian-Romanian. Parallel data is available only for Italian-English, Romanian-English, German-English, and Dutch-English. We leverage multi-lingual neural machine translation trained on all available parallel data to translate across Italian-Romanian (in both directions), either directly (zero-shot) or through English (pivoting).}
\label{figure:scenario}
\end{figure}

Application scenarios in which zero-shot translation can bootstrap the creation of new parallel data -- \textit{e.g.} via human post-editing--~\cite{johnson2016google}, show how translation performance in the initial zero-shot direction improves over time with the addition of new parallel data. In this work, we  explore instead the possibility to enable a trained Multi-NMT model to further learn from its own generated data. Briefly, our method works as follows: first (1), we let the Multi-NMT engine generate zero-shot translations on some portion of the training data; then (2), we re-start the training process on both the generated translations and the original parallel data. We repeat this \textit{training-inference-training} cycle for a few times. Notice that, at each iteration, the original training data is augmented only with the last batch of generated translations. We observe that the generated outputs initially contain a mix of words from the shared vocabulary, but after few iteration they tend to only contain words in the zero-shot target language thus becoming more and more suitable for learning. 

We test our approach on a Multi-NMT scenario including Italian, Romanian, English, German and Dutch, assuming that the zero-shot translation pair is Italian-Romanian. We also make the assumption that all languages 
have just parallel data with English (see Figure \ref{figure:scenario}). We apply our approach on top of the multilingual NMT training method suggested by~\cite{johnson2016google}. Experimental results show that our iterative training procedure not only significantly improves performance on the zero-shot directions, but it also boost multilingual NMT in general. Finally, our approach shows to outperform pivot-based machine translation, too.

\section{Related Work}
\label{related}
In this section we discuss relevant works on multilingual NMT, zero-shot NMT, and model training with self-generated data, which are closely related to our approach.

\subsection{Multilingual NMT}
Previous works in Multi-NMT are characterized by the use of separate encoding and/or decoding networks for every translation direction. Dong et al. (2015)~\cite{dong2015multi} proposed  a multi-task learning approach for a \textit{one-to-many} translation scenario, based on a sharing representations between related tasks --\textit{i.e} the source language -- in order to enhance  generalization on the target language. In particular, 
they used a single encoder in the source side, and separate attention mechanism and decoders for every target language. In a related work~\cite{luong2015multi}, used separate encoder and decoder networks for modeling language pairs in a \textit{many-to-many} setting. Notably, they dropped the attention mechanism
in favor of a shared vector space where to represent both text and multi-modal information. Aimed at reducing ambiguities at translation time,~\cite{zoph2016multi} employed a \textit{multi-source} system that considers two languages on the encoder side and one target language on the decoder side. In particular, the attention model is applied to a combination of the two encoder states. In a \textit{many-to-many} translation scenario,~\cite{firat2016multi} introduced a way to share the attention mechanism across multiple languages. As in~\cite{dong2015multi}, but (\textit{only on the decoder side}) and in~\cite{luong2015multi}, they used separate encoders and decoders for each source and target language.

Despite the reported improvements, the need of using additional encoder and/or decoder for every language added to the system tells the limitation of these approaches, by making their network complex and expensive to train. 

In a very different way,~\cite{johnson2016google} and~\cite{ha2016toward} developed similar Multi-NMT approaches by introducing a \textit{target-forcing} token in the input. The approach in~\cite{ha2016toward} applies a language-specific code to words from different languages in a mixed-language vocabulary. 
In practice, they force the decoder to translate to a specific target language by prepending and appending an artificial token to the source text. However, their word and sub-word level language-specific coding mechanism significantly increase the input length, which shows to have an impact on the computational cost and performance of NMT~\cite{cho2014properties}. In~\cite{johnson2016google}, only one artificial token is prepended to the source sentences in order to specify the target language. Prepending language tokens has permitted to eliminate the need of having separate encoder/decoder networks and attention mechanism for every new language pair.

\subsection{Zero-Shot Translation} 
By extending the approach in~\cite{firat2016multi}, zero-resource NMT has been suggested in~\cite{firat2016zero}. The authors proposed a \textit{many-to-one} translation setting and used the idea of generating a pseudo-parallel corpus~\cite{sennrich2015improvingMono}, using a pivot language, to fine tune their model. Moreover, also in this case the need of separate encoders and decoders for every language pair 
significantly increases the complexity of the model.

An attractive feature of the \textit{target-forcing} mechanism  comes from the possibility to perform zero-shot translation with the same multilingual setting as in~\cite{johnson2016google,ha2016toward}. However, recent experiments have shown that the mechanism fails to achieve reasonable zero-shot translation performance for low-resource languages \cite{MNMTlow-resourceSurafel}. The promising results in~\cite{johnson2016google} and~\cite{ha2016toward} hence require further investigation to verify if their method can work in various language settings, particularly across distant languages.

\subsection{Training with self-generated data} 
Training procedures using self-generated data have been around for a while. For instance, in statistical machine translation (SMT),~\cite{oflazer2007exploringIncremental,bechara2011statisticalIncremental} showed how the output of a translation model can be used iteratively to improve results in a task like post-editing. Mechanisms like back-translating the target side of a single language pair have been used for domain adaptation~\cite{bertoldi2009domain} and more recently by~\cite{sennrich2015improvingMono} to improve an NMT baseline model. In~\cite{dual-learningMT}, a dual-learning mechanism is proposed where two NMT models working in the opposite directions provide each other feedback signals that permit them to learn from monolingual data. In a related way, our approach also considers training from monolingual data along dual zero-shot directions. As a difference, however, our \textit{train-infer-train} loop leverages the capability of the network to jointly learn multiple translation directions.

Although our brief survey shows that re-using the output of an MT system for further training and improvement has been successfully applied in different settings, our approach differs from past works in mainly two aspects: \textit{i)} introducing for the first time a \textit{train-infer-train} mechanism addresses Multi-NMT, and \textit{ii)} we cast the approach into a \textit{self-correcting} training procedure over two dual zero-shot directions, so that incrementally improved translations mutually reinforce each direction.

\section{Neural Machine Translation}
\label{sec:nmt}
\begin{figure}[!t]
\centerline{\epsfig{figure=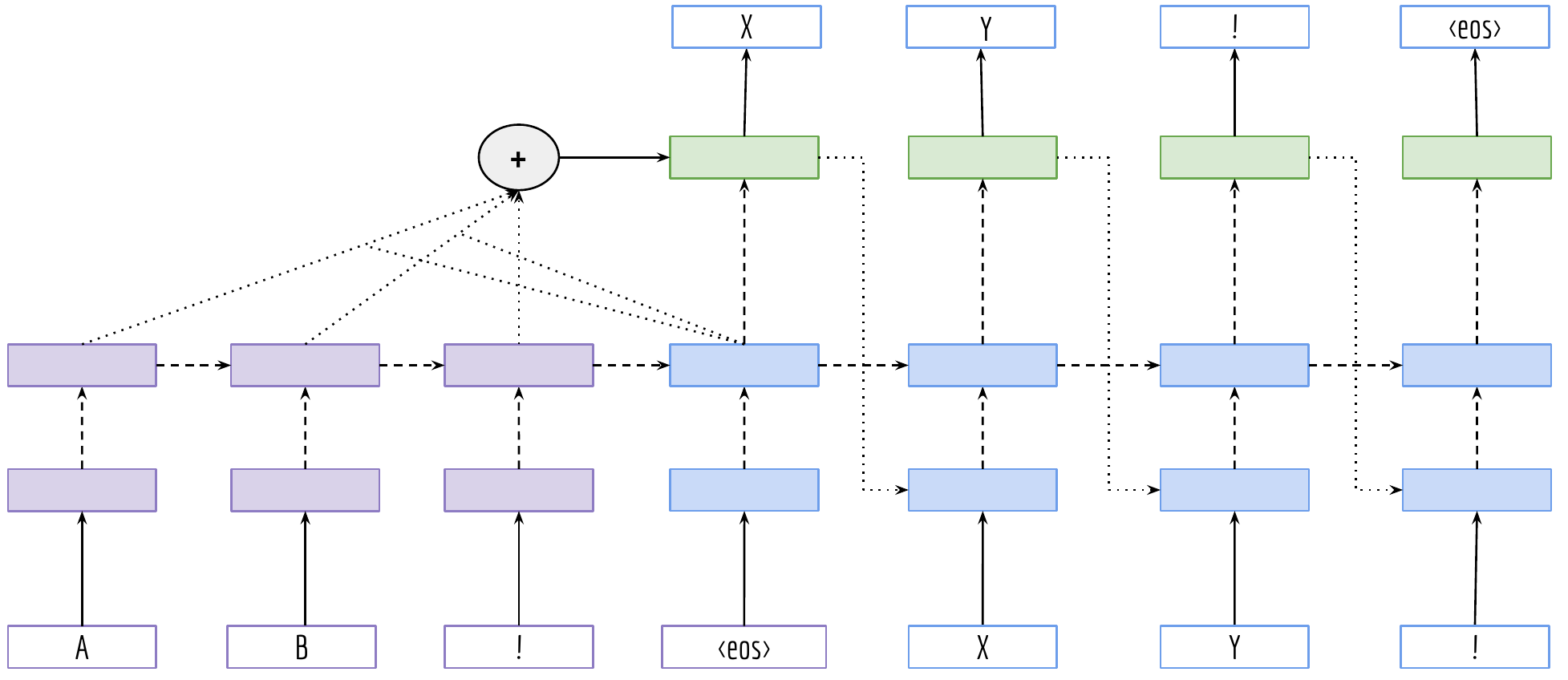,width=3.4in}}
\caption{{\it NMT architecture with encoder-decoder and an attention mechanism, showing a source sentence ``A B'' translated into a target sentence ``X Y''.}
\label{figure:encoder-decoder}}
\end{figure}

The standard NMT architecture comprises an encoder, a decoder and an attention-mechanism, which are all trained  with maximum likelihood in an end-to-end fashion~\cite{bahdanau2014neural}. The encoder is a  recurrent neural network (RNN) that encodes a source sentence into a sequence of hidden state vectors. The decoder is another RNN that uses the representation of the encoder to predict words in the target language~\cite{cho2014properties}~\cite{cho2014learningGRU}. As the name suggests, \textit{attention} is a mechanism used to improve the translation quality by deciding which part of the source sentence can contribute mostly in the prediction process \cite{luong2015effective}. As shown in Figure \ref{figure:encoder-decoder}, which simplifies the NMT architecture, first the encoder takes the source  words on the left (purple color), maps them to vectors and feeds them into the RNN. When the $<$eos$>$ (\textit{i.e} end of sentence) symbol is seen, the final time step initializes the decoder RNN (blue color).  At each time step, the attention mechanism is applied over the encoder hidden states and combined with the current hidden state of the decoder to predict the next target word. Then, the prediction is fed back to the encoder RNN to predict the next word, until the $<$eos$>$ symbol is generated~\cite{klein2017opennmt}.

\section{Mixed Language Input for Multi-NMT}
\label{sec:method}
\begin{figure*}[!t]
\centerline{\epsfig{figure=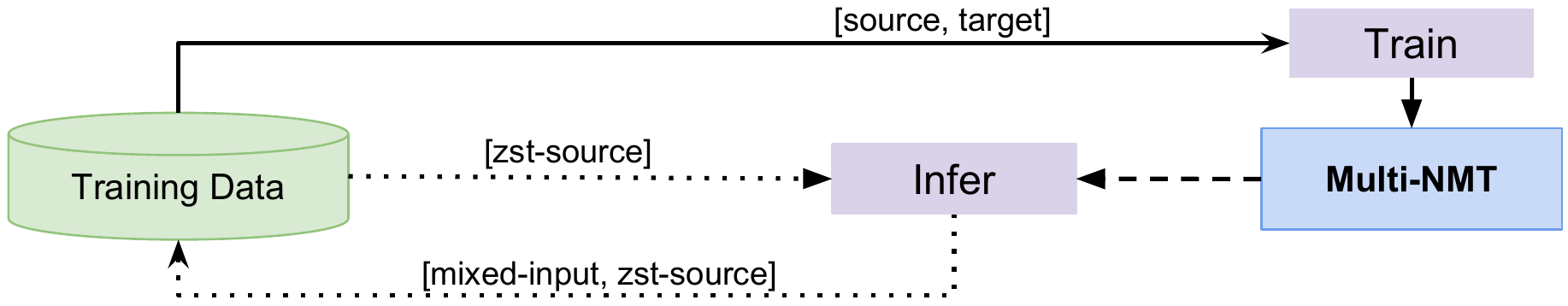,width=5.5in}}
\caption{{\it Illustration of the proposed multilingual \emph{train-infer-train} strategy. Using a standard NMT architecture, a portion of two zero-shot directions monolingual dataset is extracted for inference to construct a dual source$\leftrightarrow$target mixed-input and continue the training. The top solid line shows the training process, where as the dashed lines show the inference stage}}
\label{fig:train-infer-train_illustration}
\end{figure*}

Our goal is to improve translation in the zero-shot directions of a  multilingual model with limited directions covered by the training data (see Figure 1). The training strategy of the proposed approach is summarized in Algorithm \ref{table:training_strategy}, while its flow chart is illustrated in Figure~\ref{fig:train-infer-train_illustration}. 

To address this problem, our training procedure is performed in three  steps which are iterated for several rounds. In the first step (line $2$), the multilingual NMT system is trained on the original data available. In the second step (line $5$), the trained model is run to translate between the zero-shot directions. Then, in the third step (line $8$), the output translations are combined with the corresponding source sentences and added to the original training data. The resulting expanded corpus is now ready to perform a new round  of the training process. 

According to our \textit{train-infer-train} scheme, new \textit{synthetic} data for the two zero-shot directions are generated at each round. This process creates a \emph{duality} between the two zero-shot translation directions, which we can exploit for mutual improvement. Indeed, for each direction, sub-optimal translations $t^*$ paired with the corresponding original (and well-formed) source $s$ are used to obtain new ``parallel'' ($t^*$,$s$) sentence pairs that extend the training material for the other direction. The translated mixed-input for the two languages is represented as $T^{*}$, while the target side $T$ represents the original sentences extracted for inference.

\begin{table}[!t]
\vspace{2mm}
\centerline{
\begin{tabular}{l}
\hline
\textbf{Algorithm 1:} Iterative Learning Procedure \\ 
\hline
1: TRAIN: $D$($src, tgt$) \\
2: \hspace{2mm} Multi-NMT $\leftarrow$ initial training using dataset $D$ \\
3: \textbf{repeat} INFER-TRAIN \\
4: \hspace{2mm} \textbf{for} $s$ $=$ $1$,$T$ \textbf{do} \\
5: \hspace{5mm} $t^{*}$ $\leftarrow$ inference in duality using Multi-NMT \\
6: \hspace{2mm} \textbf{end for} \\
7: \hspace{2mm} prepare $D^{*}$([$src$ + $T^{*}$], [$tgt$ + $T$])  \\
8: \hspace{2mm} Multi-NMT  $\leftarrow$ reload Multi-NMT, train using $D^{*}$ \\
9: \hspace{2mm} return Multi-NMT \\ 
10: \textbf{until} Multi-NMT converges $\rightarrow$ Multi-NMT$^*$ \\
\hline 
\end{tabular}}
\caption{\label{table:training_strategy} {\it Iterative Learning algorithm of the proposed approach using the duality of zero-shot translation directions.}}
\end{table}

In the Multi-NMT scenario, the sub-optimal translations representing the source element of the new training pairs will likely contain a mixed-language that includes words from a vocabulary shared with other languages.
The expectation is that, round after round, the model will generate better outputs by learning at the same time to translate and ``correct'' its own translations by removing spurious elements from other languages. If this intuition holds true, the iterative improvement will yield increasingly better results in  translating between the $source$$\leftrightarrow$$target$ zero-shot directions. Ideally, this incremental training and inference cycle can continue until the model converges (line $10$).

\section{Experiments}\label{experiments}
All the experiments are carried out using the open source OpenNMT-py\footnote{https://github.com/OpenNMT/OpenNMT-py} toolkit~\cite{klein2017opennmt}. For training the models, we used the parameters specified in Table \ref{table:parameters}. Considering the high data sparsity
of our low-resource setting, we applied a dropout of $0.3$~\cite{gal2016theoreticallyDropout} to prevent overfitting~\cite{srivastava2014dropout}.
To train the baseline Multi-NMT, we used Adam~\cite{kingma2014adam} as the optimization algorithm with an initial learning rate of $0.001$. In the subsequent \textit{train-infer-train} rounds, we used SGD \cite{dean2012largeSGD}, with a learning rate of $1$. If the perplexity does not decrease on the validation set or the number of epoch is above $7$, a learning rate decay of $0.7$ is applied. This combination of optimizers was found to be effective in accelerating the training in the first few iterations. 
In all the reported experiments the baseline models are trained until convergence, while each train round after the inference stage is assumed to iterate over 10 epochs. For decoding, a beam search of size 10 is applied. 

\begin{table}[!t]
\vspace{2mm}
\centerline{
\begin{tabular}{|c|c|}
\hline
Model parameters & Value \\
\hline  \hline
RNN type & LSTM \\
RNN size & 1024 \\
Embedding dim & 512  \\
Encoder & bidirectional \\
Encoder depth & 2 \\
Decoder depth & 2 \\
Beam size & 10 \\
Batch size & 128 \\
\hline
\end{tabular}}
\caption{\label{table:parameters} {\it Hyper-parameters used to train all the models, unless specified in a different setting.}}
\end{table}

\subsection{Dataset}
To evaluate our approach, we consider five languages (\textit{i.e} English (EN), Dutch (NL), German (DE), Italian (IT), and Romanian (RO)). To simulate a low-resource scenario, each language pair has $\approx200k$ parallel sentences (see Table \ref{table:datasets} for details). 
All the parallel datasets are from the IWSLT17~\footnote{https://sites.google.com/site/iwsltevaluation2017/} multilingual shared task~\cite{cettolo2012wit3}.

\begin{table}[!t]
\vspace{2mm}
\centerline{
\begin{tabular}{|c|c|c|c|}
\hline
Direction & Training & test2010 & test2017  \\
\hline  \hline
EN $\leftrightarrow$ DE & 197,489 &  1,497 & 1,138 \\
EN $\leftrightarrow$ IT & 221,688 &  1,501 & 1,147 \\
EN $\leftrightarrow$ NL & 231,669 &  1,726 & 1,181 \\
EN $\leftrightarrow$ RO & 211,508 & 1,633 & 1,129 \\
\hline
IT $\leftrightarrow$ RO & 209,668 & 1,605 & 1,127 \\
\hline
\end{tabular}}
\caption{\label{table:datasets} {\it Number of sentences  used to train the multilingual model on eight directions. The IT $\leftrightarrow$ RO pairs are used to train only the bilingual models.}}
\end{table}

\noindent
To train all models, we used the same pipeline, first to get a tokenized dataset. Then, we apply byte pair encoding (BPE)~\cite{sennrich2015sub-word}, using a jointly trained (on source and target dataset) shared BPE model to segment the tokens into sub-word units. For this operation we used $8,000$ BPE merging rules, with a minimum of $30$ frequency threshold to apply the segmentation. When training the multilingual models, the pipeline includes adding the artificial language token at the source side of  each parallel dataset both for the training and validation sets \cite{johnson2016google}. We evaluate our models using test2010, and for comparison we use test2017 of the IWSLT2017 evaluation dataset.

\subsection{Models}
Our baseline models are trained in a multilingual and bilingual settings. For each direction of the multilingual model and every bilingual model we report the BLEU \cite{papineni2002bleu} score computed using \emph{multi-bleu.perl}~\footnote{http://www.statmt.org/moses} from the Moses SMT implementation.
BLEU scores of the Multi-NMT systems trained on the parallel data in Table $\ref{table:datasets}$ are reported in Table $\ref{8-direction}$ and $\ref{8-direction2}$ (second column).
To compare our zero-shot translations against those of the bilingual models we trained two Italian$\leftrightarrow$Romanian models. Both bilingual are trained with the same amount of training data
used by each direction of the Multi-NMT model (see Table~\ref{table:datasets}).
Moreover, as additional terms of comparison, we trained two pivoting-based systems (using English as a pivot language): Italian$\rightarrow$English$\rightarrow$Romanian and Romanian$\rightarrow$English$\rightarrow$Italian.

\subsubsection{Bilingual models}
The baseline models for comparison consist of: \textit{i)} an eight direction multilingual model (Multi-NMT), and two bilingual NMT models. 

\begin{table}[!t]
\begin{center}
\begin{tabular}{|c|c|c|c|}
\hline \bf System & \bf tst2010 & \bf tst2017\\ \hline
\hline
Italian$\rightarrow$Romanian & 19.66 & 19.14 \\ \hline
Romanian$\rightarrow$Italian & 22.44 & 20.69  \\ \hline
\end{tabular}
\end{center}
\caption{BLEU scores of two bilingual NMT models (Italian$\rightarrow$Romanian and Romanian$\rightarrow$Italian) on IWSLT data \textit{tst2010} and \textit{tst2017}}
\label{bi-nmt}
\end{table}

\noindent
The results of the two bilingual models are shown in Table~\ref{bi-nmt}. From the Multi-NMT model (see row $9$ and $10$ of Table~\ref{8-direction} and Table~\ref{8-direction2}), we particularly focus on the performance of the zero-shot directions that can be compared with the results from these two models.

\subsubsection{Pivoting} 
If data are available, the pivoting strategy is the most intuitive way to accomplish zero-shot translation, or to translate from/into under-resourced languages through high resource ones~\cite{wu2007pivot}. However, results in the pivoting framework are strictly bounded to the performance of the two combined translation engines, and especially to that of the weaker one. In contrast, Multi-NMT models that leverage knowledge acquired from data for different language combinations (similar to multi-task learning) can potentially compete or even outperform the pivoting ones. Checking  this possibility is the motivation for our comparison between the two approaches.

\begin{table}[!t]
\begin{center}
\begin{tabular}{|c|c|c|c|}
\hline \bf System & \bf tst2010 & \bf tst2017\\ \hline
\hline
Italian$\rightarrow$Romanian & 16.4 & 15.00 \\ \hline
Romanian$\rightarrow$Italian & 18.9 & 17.36 \\ \hline
\end{tabular}
\end{center}
\caption{Performance of the Italian$\leftrightarrow$Romanian pivot translation directions using English as a pivot on \textit{tst2010} and \textit{tst2017}}
\label{pm-nmt}
\end{table}

In our experiment we take English as the bridge language between Italian and Romanian in both translation directions. Unsurprisingly, compared with those of the bilingual models trained on Italian$\leftrightarrow$Romanian data, the results shown in Table \ref{pm-nmt} are lower.

On both translation directions, the bilingual models are indeed about $3.0$ BLEU points better. Such comparison, however, is not the main point of our experiment, instead, we aim to fairly analyze performance differences between pivoting and zero-shot methods trained in the same condition which lacks Italian$\leftrightarrow$Romanian training data.

\subsection{Zero-shot results}
In this experiment, we show how our approach helps to improve the baseline Multi-NMT model.  
The \textit{train-infer-train} procedure described in Section~\ref{sec:method} was applied
for \sml{five} rounds, where each round consists of $10$ iterations.  Table \ref{8-direction}, shows the improvement on the Italian$\leftrightarrow$Romanian zero-shot directions using the Multi-NMT$^*$ model. Specifically, the Italian$\rightarrow$Romanian direction reached $17.38$ BLEU score improving over the baseline ($8.59$) by $8.79$ points. Romanian$\rightarrow$Italian translation improved with an even larger margin (+$10.71$) from $8.65$ to $19.36$ BLEU score.  \\ 

\begin{table}[!t]
\begin{center}
\begin{tabular}{|l|c|c|c|}
\hline 
\cline{2-3} & \bf Multi-NMT & \bf Multi-NMT$^*$ \\ \hline
\hline
English$\rightarrow$Italian & 27.07   & \textbf{28.47} \\ \hline
Italian$\rightarrow$English & 32.12   & \textbf{33.16} \\ \hline
English$\rightarrow$Romanian & 24.65  & \textbf{25.37} \\ \hline
Romanian$\rightarrow$English & 32.7    & \textbf{34.00} \\ \hline
English$\rightarrow$German & 26.39    & \textbf{26.42} \\ \hline
German$\rightarrow$English & 31.3    & \textbf{31.79} \\ \hline
English$\rightarrow$Dutch & 30.27   & \textbf{30.85} \\ \hline
Dutch$\rightarrow$English & 35.13    & \textbf{35.77} \\ \hline
\hline
Italian$\rightarrow$Romanian & 8.59  & \bf{17.38} \\ \hline
Romanian$\rightarrow$Italian & 8.65  & \bf{19.36} \\ \hline
\end{tabular}
\end{center}
\caption{\label{8-direction} Comparison on \textit{test2010} set between a baseline Multi-NMT model against a Multi-NMT$^*$ model with our proposed  \textit{train-infer-train} approach for the Italian$\leftrightarrow$Romanian zero-shot direction. The best result for each direction is shown in bold.}
\end{table}

\begin{table}[!t]
\begin{center}
\begin{tabular}{|l|c|c|}
\hline 
\cline{2-3} & \bf Multi-NMT & \bf Multi-NMT$^*$  \\\hline
\hline
English$\rightarrow$Italian &  29.02 & \textbf{30.43}  \\ \hline
Italian$\rightarrow$English &  32.87 & \textbf{33.61} \\ \hline
English$\rightarrow$Romanian & 20.96 & \textbf{21.94} \\ \hline
Romanian$\rightarrow$English & 27.48 & \textbf{28.21} \\ \hline
English$\rightarrow$German &  19.75  & \textbf{19.85} \\ \hline
German$\rightarrow$English &  24.12  & \textbf{24.25} \\ \hline
English$\rightarrow$Dutch &  25.37 &   \textbf{26.12} \\ \hline
Dutch$\rightarrow$English &  \textbf{29.25} &   29.15 \\ \hline
\hline
Italian$\rightarrow$Romanian & 8.18 &  \textbf{17.08} \\ \hline
Romanian$\rightarrow$Italian & 8.58 &  \textbf{19.25} \\ \hline
\end{tabular}
\end{center}
\caption{\label{8-direction2} Comparison on \textit{test2017} set between a baseline Multi-NMT model against a Multi-NMT$^*$ model with our proposed  \textit{train-infer-train} approach for the Italian$\leftrightarrow$Romanian zero-shot direction. 
}
\end{table}

\noindent
In addition, and to our great surprise, the results from our self-correcting mechanism showed to perform even better than the pivoting strategy. To check the validity of our results, we also compared the baseline multilingual system and our approach on the IWSLT 2017 test set (\textit{test2017}). As shown in Table \ref{8-direction2}, the results confirm those computed on \textit{test2010}, with almost identical gains (+$8.9$ and +$10.67$). 

The other important advantage of our approach is evidenced by the performance gains obtained on the language directions supported by  parallel training corpora. To put this into perspective, all translation directions have shown improvements, except for the slight drop (-$0.10$ BLEU) observed for the Dutch$\rightarrow$English direction in $\textit{test2017}$ case.

\begin{figure}[!htbp]
\centerline{\epsfig{figure=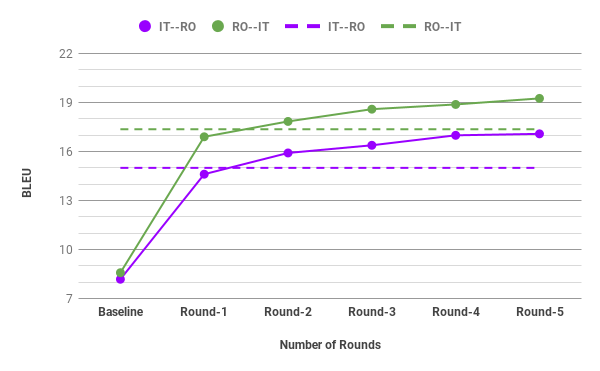,width=3.6in}}
\caption{{\it Results from test2017 for the Italian$\leftrightarrow$Romanian zero-shot directions, comparing our iterative learning approach (solid lines) with the pivoting mechanism (dashed lines)}
\label{fig:pivot-vs-iterative}}
\end{figure}

\noindent
Comparing the results from every rounds (see Figure \ref{fig:pivot-vs-iterative}), we observe that the training after the first inference step is responsible for the largest portion of the overall gain. This is mainly due to the initial introduction of (noisy) parallel data for the zero-shot directions. The contribution of the self-correcting process can be seen in the following rounds, i.e the improvement after each inference suggests that the generated data are getting cleaner and cleaner. 

\begin{table*}[!t]
  \centering
  \begin{tabular}{ll}
    \hline
      & Italian$\rightarrow$Romanian \\ \hline
    \textbf{Source} & ... che rafforza la corruzione, l'evasione fiscale, la povertà, l'instabilità. \\
    Pivot & ... poarta de bază, evazia fiscală, sărăcia, instabilitatea. \\
    Multi-NMT & ... restrânge corrupția, \textbf{fiscale de evasion, poverty, instabilitate}. \\
    Multi-NMT$^*$ & ... care rafinează corupția, \textbf{evasarea fiscală, sărăcia, instabilitatea}. \\
    \textbf{Reference} & ... care protejează corupţia, evaziunea fiscală, sărăcia şi instabilitatea. \\
     \hline
     & Romanian$\rightarrow$Italian \\ \hline
    \textbf{Source} & E o poveste incredibilă. \\
    Pivot & È una storia incredibile \\
    Multi-NMT & È una storia \textbf{incredible}. \\ 
    Multi-NMT$^*$ & È una storia \textbf{incredibile} \\
    \textbf{Reference} & È una storia incredibile . \\
	\hline
     & English$\rightarrow$Italian \\ \hline
    \textbf{Source} & We can't use them to make simple images of things out in the Universe. \\
    Multi-NMT & Non possiamo usarli \textbf{per creare immagini semplici di} cose nell'universo.  \\
    Multi-NMT$^*$ &  Non possiamo usarle \textbf{per fare semplici immagini di} cose nell'universo. \\
    \textbf{Reference} &  Non possiamo usarle per fare semplici immagini di cose nell'univero \\ \hline    
  \end{tabular}
  \caption{Top two examples: zero-shot translations generated by pivoting via English, multilingual translation(Multi-NMT) and multilingual translation enhanced with out approach (Multi-NMT$^*$). Last example: multilingual and enhanced multi-lingual translation in a resourced translation direction. }
  \label{table:sample-translation}
\end{table*}

Looking at the sample translation outputs using the different approaches in Table \ref{table:sample-translation}, we observe that the baseline Multi-NMT system produces mixed language output (\textit{e.g.} ``poverty'' in Italian$\rightarrow$Romanian and ``incredible'' in Romanian$\rightarrow$Italian). Thanks to our approach,  the Multi-NMT$^*$ system instead tends to produce more consistent target language (``poverty'' becomes ``sărăcia'' in Italian$\rightarrow$Romanian and ``incredible'' becomes ``incredibile'' in Romanian$\rightarrow$Italian). Furthermore, even in the non-zero-shot directions there are case where the enhanced Multi-NMT$^*$ system produces better translations (see the last reported example).

\section{Conclusions} 
We introduced a method to improve zero-shot translation in multilingal NMT under particularly resource-scarce training conditions. The proposed self-correcting  procedure, by leveraging synthetic dual translations, achieved significant improvements over a multilingual NMT baseline and outperformed a pivoting NMT approach for the Italian-Romanian directions. 

In future work, we plan to improve the train-infer-train stages to reach better performance in less time and with lower training complexity. In our current setup we did not consider any form of selection for the dataset to be translated at the inference stage of the train-infer-train procedure. We expect that applying frequency and similarity based approaches to select promising training candidates can bring further improvements. Moreover, we plan also to test our approach with additional monolingual data from the two zero-shot directions. Finally we plan to extend our approach to the translation of mixed language sentences (i.e code-mixing).

\section{Acknowledgements}
This work has been partially supported by the EC-funded projects ModernMT (H2020 grant agreement no. 645487) and QT21 (H2020 grant agreement no. 645452). This work was also supported by The Alan Turing Institute under the EPSRC grant EP/N510129/1 and by a donation of Azure credits by Microsoft. We gratefully acknowledge the support of NVIDIA Corporation with the donation of the Titan Xp GPU used for this research.

\bibliographystyle{IEEEtran}
\bibliography{bib}

\begin{thebibliography}{10}
\providecommand{\url}[1]{#1}
\csname url@rmstyle\endcsname
\providecommand{\newblock}{\relax}
\providecommand{\bibinfo}[2]{#2}
\providecommand\BIBentrySTDinterwordspacing{\spaceskip=0pt\relax}
\providecommand\BIBentryALTinterwordstretchfactor{4}
\providecommand\BIBentryALTinterwordspacing{\spaceskip=\fontdimen2\font plus
\BIBentryALTinterwordstretchfactor\fontdimen3\font minus
  \fontdimen4\font\relax}
\providecommand\BIBforeignlanguage[2]{{%
\expandafter\ifx\csname l@#1\endcsname\relax
\typeout{** WARNING: IEEEtran.bst: No hyphenation pattern has been}%
\typeout{** loaded for the language `#1'. Using the pattern for}%
\typeout{** the default language instead.}%
\else
\language=\csname l@#1\endcsname
\fi
#2}}

\bibitem{koehn2017six}
P.~Koehn and R.~Knowles, ``Six challenges for neural machine translation,''
  \emph{arXiv preprint arXiv:1706.03872}, 2017.

\bibitem{johnson2016google}
M.~Johnson, M.~Schuster, Q.~V. Le, M.~Krikun, Y.~Wu, Z.~Chen, N.~Thorat,
  F.~Vi{\'e}gas, M.~Wattenberg, G.~Corrado, \emph{et~al.}, ``Google's
  multilingual neural machine translation system: Enabling zero-shot
  translation,'' \emph{arXiv preprint arXiv:1611.04558}, 2016.

\bibitem{ha2016toward}
T.-L. Ha, J.~Niehues, and A.~Waibel, ``Toward multilingual neural machine
  translation with universal encoder and decoder,'' \emph{arXiv preprint
  arXiv:1611.04798}, 2016.

\bibitem{dong2015multi}
D.~Dong, H.~Wu, W.~He, D.~Yu, and H.~Wang, ``Multi-task learning for multiple
  language translation.'' in \emph{ACL (1)}, 2015, pp. 1723--1732.

\bibitem{luong2015multi}
M.-T. Luong, Q.~V. Le, I.~Sutskever, O.~Vinyals, and L.~Kaiser, ``Multi-task
  sequence to sequence learning,'' \emph{arXiv preprint arXiv:1511.06114},
  2015.

\bibitem{zoph2016multi}
B.~Zoph and K.~Knight, ``Multi-source neural translation,'' \emph{arXiv
  preprint arXiv:1601.00710}, 2016.

\bibitem{firat2016multi}
O.~Firat, K.~Cho, and Y.~Bengio, ``Multi-way, multilingual neural machine
  translation with a shared attention mechanism,'' \emph{arXiv preprint
  arXiv:1601.01073}, 2016.

\bibitem{cho2014properties}
K.~Cho, B.~Van~Merri{\"e}nboer, D.~Bahdanau, and Y.~Bengio, ``On the properties
  of neural machine translation: Encoder-decoder approaches,'' \emph{arXiv
  preprint arXiv:1409.1259}, 2014.

\bibitem{firat2016zero}
O.~Firat, B.~Sankaran, Y.~Al-Onaizan, F.~T.~Y. Vural, and K.~Cho,
  ``Zero-resource translation with multi-lingual neural machine translation,''
  \emph{arXiv preprint arXiv:1606.04164}, 2016.

\bibitem{sennrich2015improvingMono}
R.~Sennrich, B.~Haddow, and A.~Birch, ``Improving neural machine translation
  models with monolingual data,'' \emph{arXiv preprint arXiv:1511.06709}, 2015.

\bibitem{MNMTlow-resourceSurafel}
S.~M. Lakew, A.~D.~G. Mattia, and F.~Marcello, ``Multilingual neural machine
  translation for low resource languages,'' in \emph{CLiC-it 2017 – 4th
  Italian Conference on Computational Linguistics, to appear}, 2017.

\bibitem{oflazer2007exploringIncremental}
K.~Oflazer and I.~D. El-Kahlout, ``Exploring different representational units
  in english-to-turkish statistical machine translation,'' in \emph{Proceedings
  of the Second Workshop on Statistical Machine Translation}.\hskip 1em plus
  0.5em minus 0.4em\relax Association for Computational Linguistics, 2007, pp.
  25--32.

\bibitem{bechara2011statisticalIncremental}
H.~B{\'e}chara, Y.~Ma, and J.~van Genabith, ``Statistical post-editing for a
  statistical mt system,'' in \emph{MT Summit}, vol.~13, 2011, pp. 308--315.

\bibitem{bertoldi2009domain}
N.~Bertoldi and M.~Federico, ``Domain adaptation for statistical machine
  translation with monolingual resources,'' in \emph{Proceedings of the fourth
  workshop on statistical machine translation}.\hskip 1em plus 0.5em minus
  0.4em\relax Association for Computational Linguistics, 2009, pp. 182--189.

\bibitem{dual-learningMT}
Y.~Xia, D.~He, T.~Qin, L.~Wang, N.~Yu, T.~Liu, and W.~Ma, ``Dual learning for
  machine translation,'' \emph{CoRR}, vol. abs/1611.00179, 2016.

\bibitem{bahdanau2014neural}
D.~Bahdanau, K.~Cho, and Y.~Bengio, ``Neural machine translation by jointly
  learning to align and translate,'' \emph{arXiv preprint arXiv:1409.0473},
  2014.

\bibitem{cho2014learningGRU}
K.~Cho, B.~Van~Merri{\"e}nboer, C.~Gulcehre, D.~Bahdanau, F.~Bougares,
  H.~Schwenk, and Y.~Bengio, ``Learning phrase representations using rnn
  encoder-decoder for statistical machine translation,'' \emph{arXiv preprint
  arXiv:1406.1078}, 2014.

\bibitem{luong2015effective}
M.-T. Luong, H.~Pham, and C.~D. Manning, ``Effective approaches to
  attention-based neural machine translation,'' \emph{arXiv preprint
  arXiv:1508.04025}, 2015.

\bibitem{klein2017opennmt}
G.~Klein, Y.~Kim, Y.~Deng, J.~Senellart, and A.~M. Rush, ``Opennmt: Open-source
  toolkit for neural machine translation,'' \emph{arXiv preprint
  arXiv:1701.02810}, 2017.

\bibitem{gal2016theoreticallyDropout}
Y.~Gal and Z.~Ghahramani, ``A theoretically grounded application of dropout in
  recurrent neural networks,'' in \emph{Advances in neural information
  processing systems}, 2016, pp. 1019--1027.

\bibitem{srivastava2014dropout}
N.~Srivastava, G.~E. Hinton, A.~Krizhevsky, I.~Sutskever, and R.~Salakhutdinov,
  ``Dropout: a simple way to prevent neural networks from overfitting.''
  \emph{Journal of machine learning research}, vol.~15, no.~1, pp. 1929--1958,
  2014.

\bibitem{kingma2014adam}
D.~Kingma and J.~Ba, ``Adam: A method for stochastic optimization,''
  \emph{arXiv preprint arXiv:1412.6980}, 2014.

\bibitem{dean2012largeSGD}
J.~Dean, G.~Corrado, R.~Monga, K.~Chen, M.~Devin, M.~Mao, A.~Senior, P.~Tucker,
  K.~Yang, Q.~V. Le, \emph{et~al.}, ``Large scale distributed deep networks,''
  in \emph{Advances in neural information processing systems}, 2012, pp.
  1223--1231.

\bibitem{cettolo2012wit3}
M.~Cettolo, C.~Girardi, and M.~Federico, ``Wit3: Web inventory of transcribed
  and translated talks,'' in \emph{Proceedings of the 16th Conference of the
  European Association for Machine Translation (EAMT)}, vol. 261, 2012, p. 268.

\bibitem{sennrich2015sub-word}
R.~Sennrich, B.~Haddow, and A.~Birch, ``Neural machine translation of rare
  words with subword units,'' \emph{arXiv preprint arXiv:1508.07909}, 2015.

\bibitem{papineni2002bleu}
K.~Papineni, S.~Roukos, T.~Ward, and W.-J. Zhu, ``Bleu: a method for automatic
  evaluation of machine translation,'' in \emph{Proceedings of the 40th annual
  meeting on association for computational linguistics}.\hskip 1em plus 0.5em
  minus 0.4em\relax Association for Computational Linguistics, 2002, pp.
  311--318.

\bibitem{wu2007pivot}
H.~Wu and H.~Wang, ``Pivot language approach for phrase-based statistical
  machine translation,'' \emph{Machine Translation}, vol.~21, no.~3, pp.
  165--181, 2007.

\end{thebibliography}

\end{document}